\documentclass[letterpaper, 10 pt, conference]{ieeeconf}  

\IEEEoverridecommandlockouts                              

\usepackage{graphicx}
\usepackage{epsfig} 
\usepackage{mathptmx} 
\usepackage{times} 
\usepackage{amsmath} 
\usepackage{amssymb}  

\usepackage{subcaption}
\usepackage{multirow}
\usepackage{hyperref}
\usepackage{xcolor}
\usepackage{cite}
\usepackage{colortbl}
\usepackage{dirtytalk}
\usepackage[normalem]{ulem}

\title{\LARGE \bf
Expanded Comprehensive Robotic Cholecystectomy Dataset (CRCD)$^{\dagger}$
}

\author{Ki-Hwan Oh$^{1*}$, Leonardo Borgioli$^{1*}$, Alberto Mangano$^{2}$, Valentina Valle$^{2}$, Marco Di Pangrazio$^{3}$,\\
Francesco Toti$^{4}$, Gioia Pozza$^{5}$, Luciano Ambrosini$^{2}$, Alvaro Ducas$^{2}$, Milo\v s \v Zefran$^{1}$, Liaohai Chen$^{2}$\\
and Pier Cristoforo Giulianotti$^{2}$
\thanks{$^{*}$ First two authors contributed equally to this work.}
\thanks{$^{\dagger}$ \url{https://github.com/sitleng/CRCD}}
\thanks{$^{1}$ Robotics Lab,  Department of Electrical and Computer Engineering, College of Engineering, University of Illinois Chicago, Chicago, IL 60607, USA.}%
\thanks{$^{2}$ Surgical Innovation and Training Lab,  Department of Surgery, College of Medicine, University of Illinois Chicago, Chicago, IL 60607, USA.}%
\thanks{$^{3}$ School of Medicine and Surgery, University of Modena and Reggio Emilia, Modena, Italy.}%
\thanks{$^{4}$ Department of Medicine and Surgery, University of Milan, Milano, Italy.}%
\thanks{$^{5}$ Department of Surgery, Lugano Regional Hospital, Ente Ospedaliero Cantonale (EOC), Lugano, Switzerland.}%
}

\begin{document}

\maketitle
\thispagestyle{empty}
\pagestyle{empty}

\begin{abstract}

In recent years, the application of machine learning to minimally invasive surgery (MIS) has attracted considerable interest. Datasets are critical to the use of such techniques. This paper presents a unique dataset recorded during ex vivo pseudo-cholecystectomy procedures on pig livers using the da Vinci Research Kit (dVRK). Unlike existing datasets, it addresses a critical gap by providing comprehensive kinematic data, recordings of all pedal inputs, and offers a time-stamped record of the endoscope's movements. This expanded version also includes segmentation and keypoint annotations of images, enhancing its utility for computer vision applications.

Contributed by seven surgeons with varied backgrounds and experience levels that are provided as a part of this expanded version, the dataset is an important new resource for surgical robotics research. It enables the development of advanced methods for evaluating surgeon skills, tools for providing better context awareness, and automation of surgical tasks. Our work overcomes the limitations of incomplete recordings and imprecise kinematic data found in other datasets. To demonstrate the potential of the dataset for advancing automation in surgical robotics, we introduce two models that predict clutch usage and camera activation, a 3D scene reconstruction example, and the results from our keypoint and segmentation models.

\end{abstract}

\section*{Acknowledgement}

Preprint of an article accepted in Journal of Medical Robotics Research (2024) \copyright\ World Scientific Publishing Company \url{https://www.worldscientific.com/worldscinet/jmrr}

\section{Introduction}

Training of state-of-the-art machine learning models requires extensive datasets. In recent years, considerable effort has been made to create public datasets for surgical procedures that include comprehensive annotations from experts. Given the potential applications of machine learning in robotic-assisted surgery (RAS), creating datasets specifically focused on robotic minimally invasive surgical tasks is of particular importance. Such datasets should provide a comprehensive characterization of the surgeon's actions as well as data describing the corresponding motion of the robot alongside the endoscopic videos.

Unfortunately, most existing datasets focus on the segmentation of instruments~\cite{bouget2015, ross2020robust} and/or organs~\cite{allan20202018, hong2020cholecseg8k, carstens2023dresden} captured by the endoscope during surgeries. For instance, Twinanda \textit{et al.}~\cite{twinanda2016endonet} created a video dataset of instrument segmentation that includes labels for different phases of the cholecystectomy procedure and used it to train a model called EndoNet for predicting instrument presence and recognizing surgical phases. However, these datasets lack the kinematic data of the robot arms and the controllers, even though kinematic data can improve instrument segmentation~\cite{su2018, da2019self} as well as improve 3D position estimation and calculation of distance from tissues.

Few state-of-the-art datasets incorporate kinematic data~\cite{dataset_review, rueckert2024methods}. For instance, Rivas-Blanco \textit{et al.}~\cite{kinematic_dataset} recorded the kinematics of controllers and surgical robot arms with external stereo cameras in fixed locations, capturing images different from endoscopes. Unlike real surgical procedures, these datasets focus on simple tasks such as moving a peg or following a wire. More advanced tasks, such as suturing and knot tying, were included in the JIGSAWS~\cite{gao2014jhu} dataset but were limited to toy experiments. Colleoni \textit{et al.}~\cite{colleoni2020synthetic} recorded kinematics to improve instrument segmentation robustness to different backgrounds, but the movements were far from practical surgical procedures.

In addition, a significant yet often overlooked set of interaction signals in existing datasets are the pedals of the robotic surgery system. Surgeons use the pedals to adjust the controllers' position (clutch), move the endoscope, and apply mono/bipolar power to dissect tissues. Analyzing these interactions and automating such secondary tasks is vital to alleviating the stress and burden on surgeons during prolonged surgical interventions.

Research evaluating surgeons' experiences with RAS has recently increased due to its growing adoption, focusing on dexterity, control, and the learning curve~\cite{hung2018comprehensive}. Further work examines the additional training required for residents transitioning from traditional laparoscopy to robotic-assisted laparoscopic surgery~\cite{yang2020training}. Hedrick \textit{et al.}~\cite{hedrick2019evaluation} used models trained on traditional laparoscopic datasets to assess surgeon performance in RAS and showed that there are common crucial skills shared between the two. Hence, the laparoscopic surgery experience is closely correlated to the RAS experience.

To address the shortcomings observed in the previously released datasets, we released the Comprehensive Robotic Cholecystectomy Dataset (CRCD)~\cite{CRCD}.
Cholecystectomy was chosen as it is one of the most popular and standard laparoscopic procedures~\cite{cullen2009ambulatory, harrell2005minimally}. The same applies to robotic cholecystectomy, which has been increasing in popularity~\cite{strosberg2017retrospective}. The original CRCD includes the following records during robotic cholecystectomy procedures: stereo endoscopic videos, kinematics data of robot arms and controllers, and pedal signals. 

In this expanded version of the dataset, we added information about the experience level of each surgeon. Moreover, we included the annotations of the pig liver segmentation and instrument keypoints based on the COCO~\cite{mscoco} format. This format is widely used in computer vision and allows new models to be easily introduced. 
Therefore, the current release of the CRCD combines qualitative information about surgeons' backgrounds, video recordings, kinematics data, pedal signals, and annotations for both tissue segmentation and instrument keypoint tracking. This unique dataset has the potential to significantly advance the field of robotic-assisted surgery by providing researchers with extensive and detailed information to improve the understanding and automation of surgical procedures.

\section{Dataset Components}

\subsection{Stereo Endoscopic Images}

The robotic platform consists of the first-generation da Vinci surgical system controlled through the dVRK~\cite{dvrk}. The da Vinci robot consists of three Patient-Side Manipulators (PSMs) and an Endoscope Camera Manipulator (ECM), but we only use two PSMs in this work. The original endoscope was replaced with the Si model~\cite{davincisi} endoscope for its superior image quality and reduced noise. Stereo endoscope cameras were calibrated using OpenCV~\cite{opencv_library} based on methods from Zhang \textit{et al.}~\cite{zhang2000} and ROS libraries~\cite{caltoolbox}, determining the intrinsic and extrinsic parameters for each camera. The dataset includes distortion parameters, intrinsic camera matrix, rectification matrix, and projection matrix for both endoscopes, enabling 3D point cloud recovery from the videos.

\begin{figure}[!ht]
    \centering
        \begin{subfigure}[b]{0.49\linewidth}
            \centering
            \includegraphics[width=\linewidth]{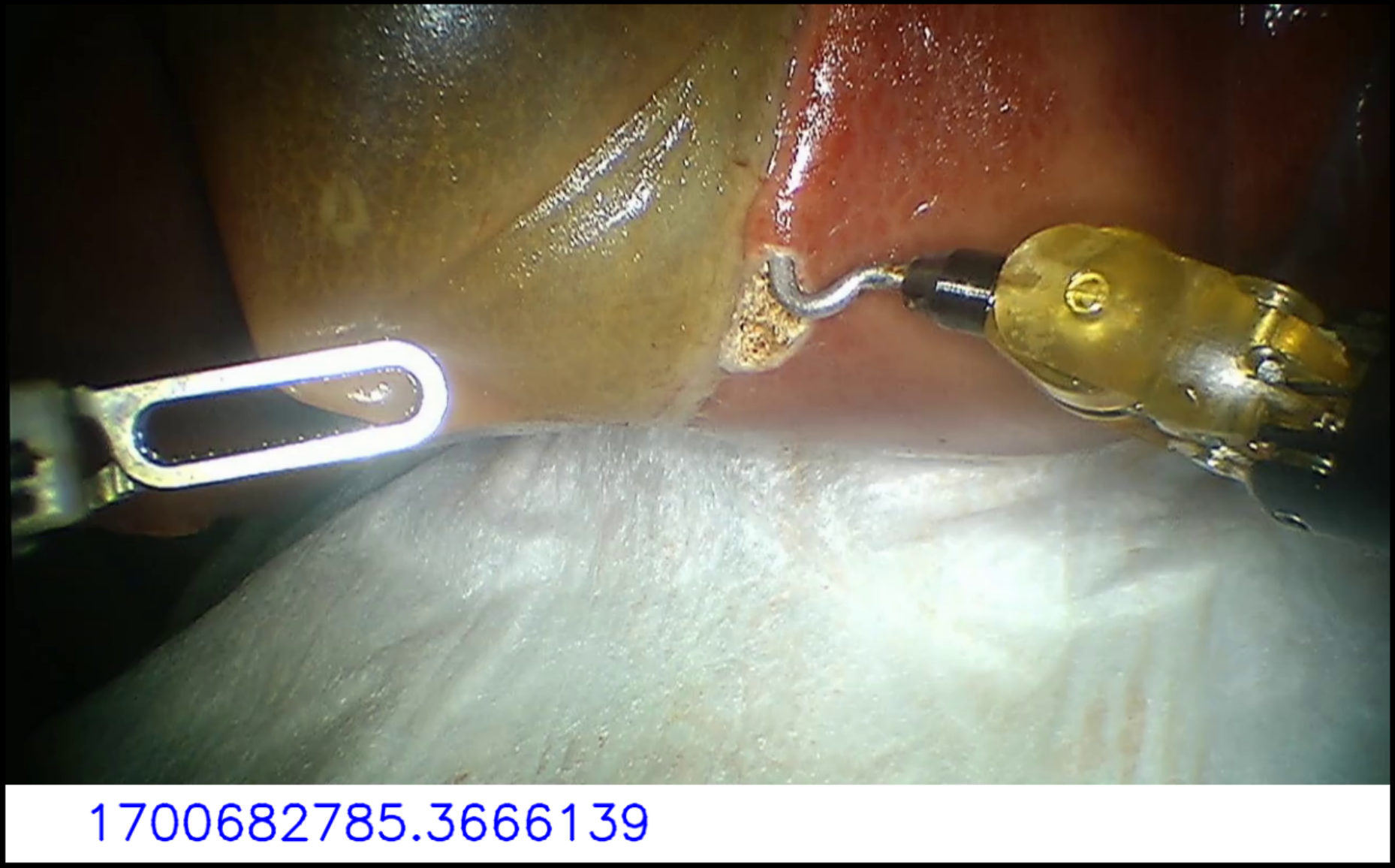}
            \caption{Left}
            \label{fig:left}
        \end{subfigure}
        \hfill
        \begin{subfigure}[b]{0.49\linewidth}
            \centering
            \includegraphics[width=\linewidth]{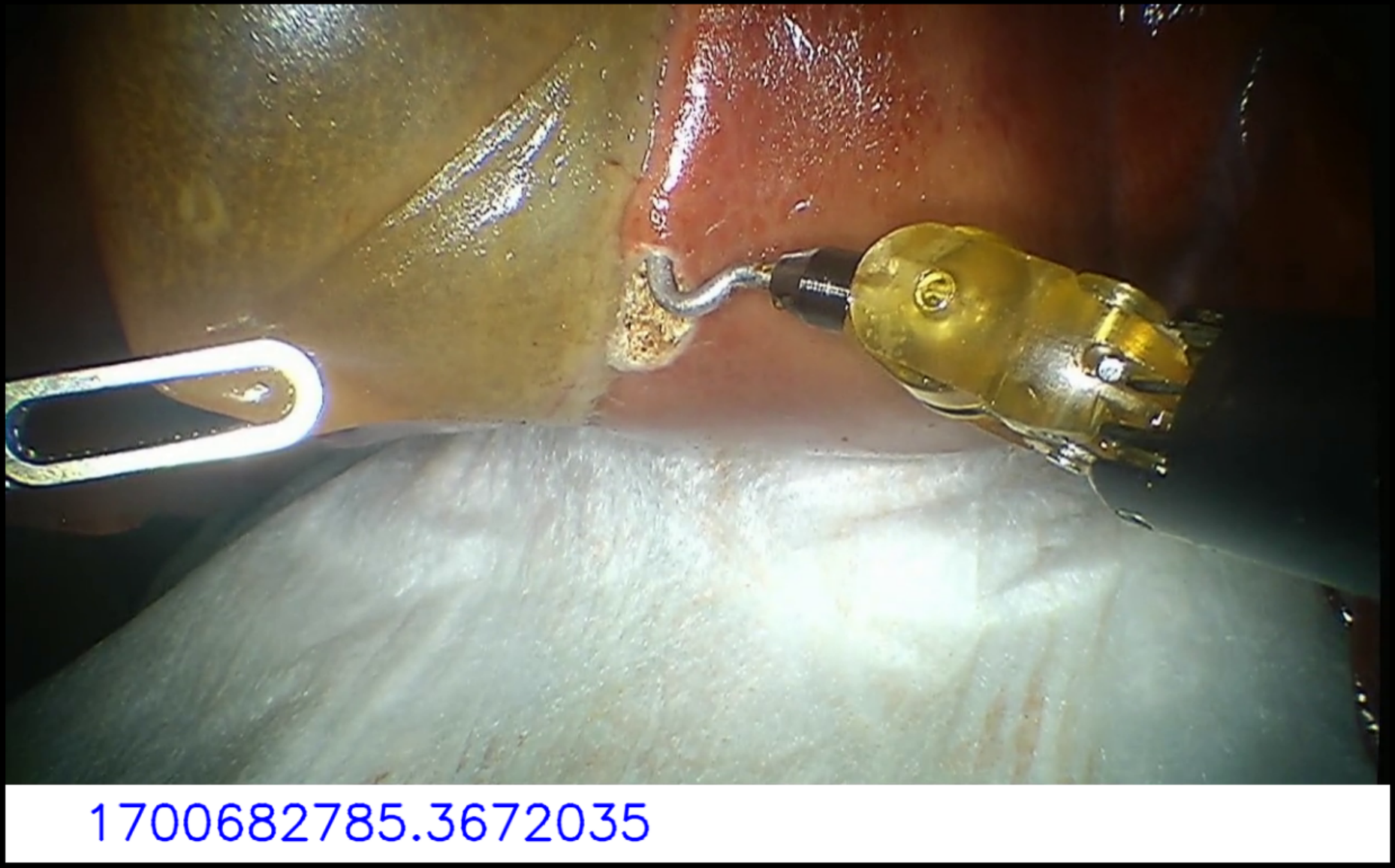}
            \caption{Right}
            \label{fig:right}
        \end{subfigure}
\caption{Sample of the stereo endoscopic images.}
\label{fig:stereo}
\end{figure}

Individual images from each camera are recorded separately with a timestamp at the bottom (Fig.~\ref{fig:stereo}). Timestamps are from the Robot Operating System (ROS)~\cite{ros} and can be extracted using OCR engines such as Tesseract~\cite{tesseract}. These timestamps link to the corresponding kinematic data and pedal signals in the dataset. Videos are recorded at 60 frames per second with a resolution of $1280 \times 720$ pixels. They are encoded in AVC1 FourCC and compressed to MP4 format.

\subsection{Pedals}

The da Vinci model includes \textit{camera}, \textit{clutch}, \textit{monopolar}, and \textit{bipolar} pedals. The dVRK provides pedal signals only when the pedals are pressed or released. To ensure synchronization with the image and kinematic data, we modified the signals to stay at $0$ by default and rise to $1$ while the corresponding pedal is pressed.

\begin{figure}[!ht]
    \centering
    \includegraphics[width=0.9\linewidth]{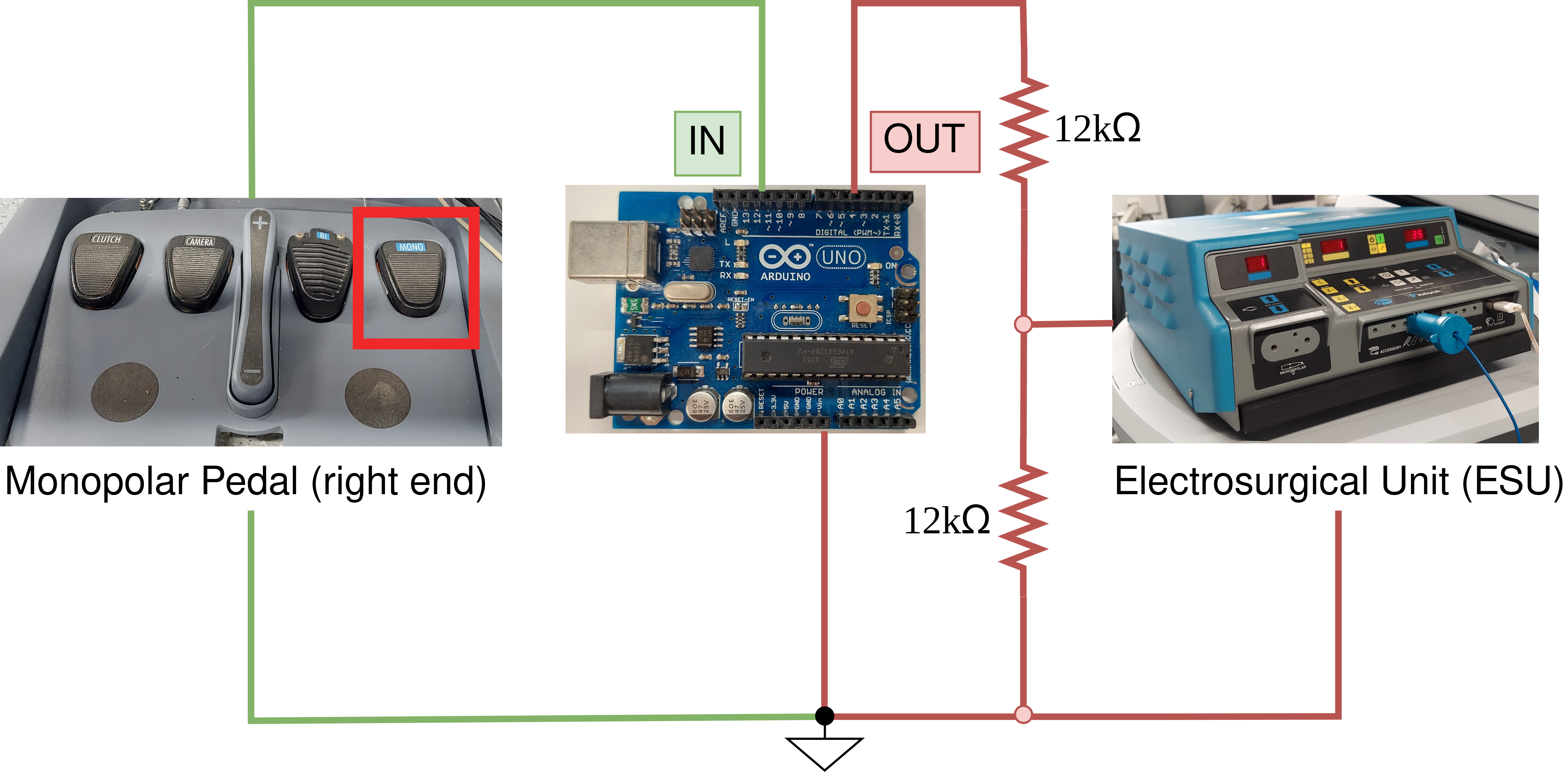}
    \caption{A schematic of a connection between the Arduino, console pedals, and the ESU.}
    \label{fig:mp_circuit}
\end{figure}

The dVRK lacks direct control over the electrosurgical unit (ESU), which controls the mono/bipolar power delivered to the instruments. We used a Pfizer Valleylab Force 2 ESU, requiring a minimum current of $1 mA$ through the input cable connected to the pedals to activate the monopolar output. We interfaced the generator's input cable with the da Vinci console pedals using an Arduino (Fig.~\ref{fig:mp_circuit}). The pedal inputs are recorded at $230$Hz.

\subsection{Kinematic Data}

The kinematic data in the dataset is based on the forward kinematics of the da Vinci robot derived from our custom calibration~\cite{oh2023framework} using fiducial markers. This calculation determines the position of the PSM tip relative to the ECM tip. 

\begin{figure}[!ht]
    \centering
    \includegraphics[width=\linewidth]{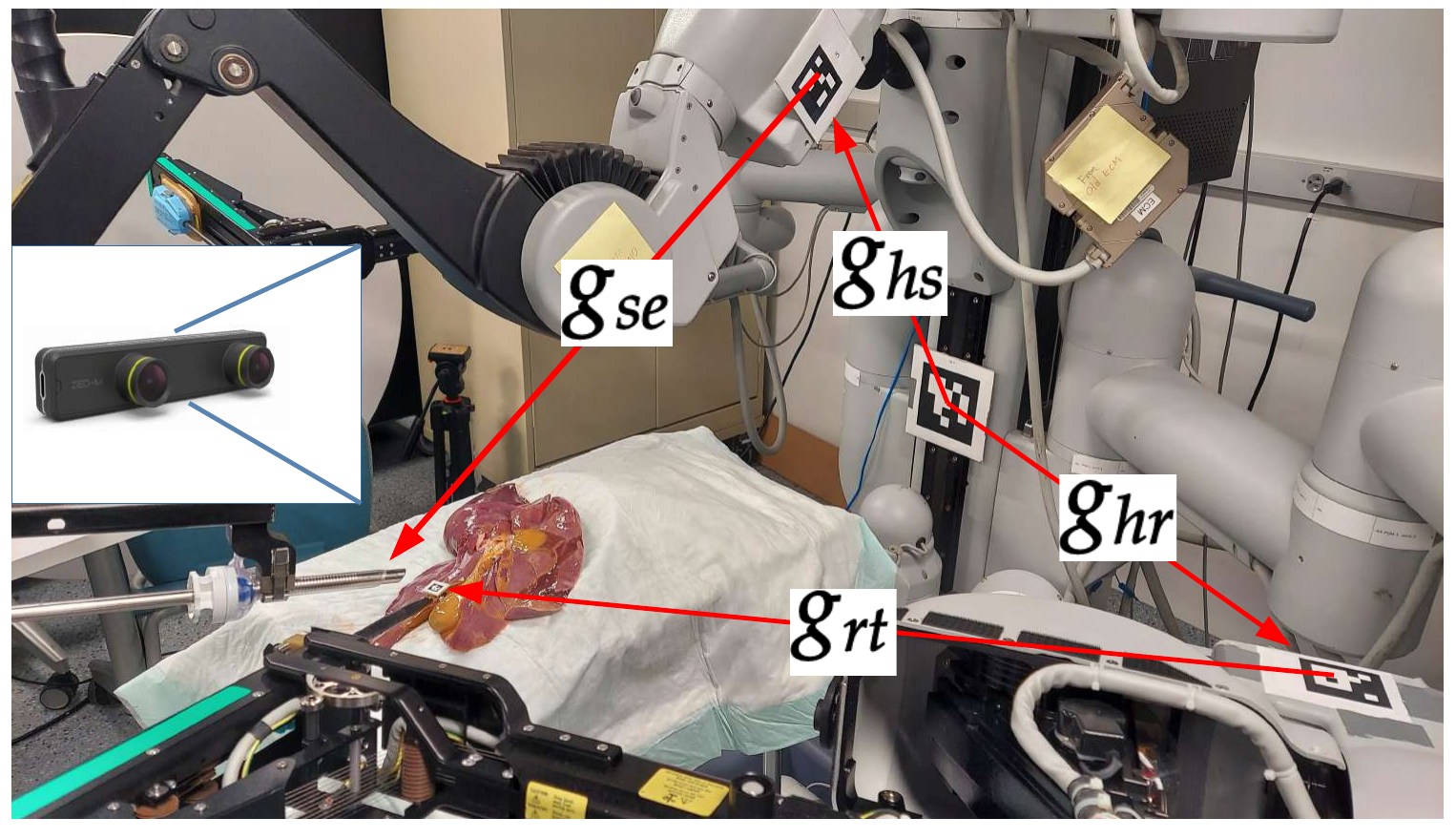}
\caption{The setup shows how our custom-calibrated kinematics work. The transformations are shown based on the direction of the arrows, and eventually, they are used to find the transformation between the ECM tip and PSM tip.}
\label{fig:calibsetup}
\vspace{-3mm}
\end{figure}

In Fig.~\ref{fig:calibsetup}, each $g_{ab}$ is the transformation (homogeneous matrix) between frames $A$ and $B$. The base frames for the PSM and ECM are $R$ and $S$, respectively, with $T$ and $E$ as their instrument tip frames, respectively. Once $g_{rt}$ and $g_{se}$ are determined, the relative configuration of the PSM tip to the ECM tip is established, incorporating the Helper ($H$) frame. If the pose of the Setup Joints (SUJs) change, only the transformation between the helper and base frames needs updating:
\begin{equation}
g_{et} = g_{se}^{-1}\cdot g_{hs}^{-1}\cdot g_{hr}\cdot g_{rt} = g_{es}\cdot g_{sh}\cdot g_{hr}\cdot g_{rt}
\end{equation}
The transformations $g_{sh}$ and $g_{hr}$ are obtained from fiducial markers using an external camera.

For the PSMs, the dataset includes the transformation from the arm's base frame to its instrument tip ($g_{rt}$), the transformation from the ECM tip to the PSM instrument tip ($g_{et}$), the joint states (position, velocity, and effort), and the jaw's joint states from the dVRK. For the ECM, the dataset includes the transformation from the arm's base frame to its instrument tip ($g_{se}$), the transformation from the Helper frame to the ECM tip ($g_{he}$), and the dVRK joint states.

For the Master Tool Manipulator (MTM), used by the surgeon to control the robot, the raw kinematic data from the dVRK~\cite{kinematic_dataset, dvrk} was recorded. This data contains the transformation from the base of each arm to its controller tip, the transformation from the High-Resolution Stereo Video (HRSV) frame (shown to the surgeon) to the controller tip, the joint states (position, velocity, and effort) of each arm, and the joint angle of each gripper~\cite{kinematic_dataset}. The PSM1 is associated with the MTMR (MTM Right), while the PSM2 is paired with MTML (MTM Left); when the camera pedal is engaged, both MTMs are linked and control the ECM. The kinematics of the robot arms and console manipulators are sampled at $100$Hz.

\subsection{Dataset for Object Instance Segmentation and Keypoint Detection}

Accurately extracting surgical instrument locations and target tissues from endoscopic images is a core challenge in automating surgical procedures. A robust dataset is essential for achieving this goal. To this end, we enhanced the CRCD's endoscopic images with instrument keypoints and image segmentation annotations. The details of this dataset can be found in Table~\ref{tab:customdata_seg}.

\begin{table}[!ht]
\centering
\resizebox{\columnwidth}{!}{%
\begin{tabular}{l|l|c|c}
\hline
\hline
\textbf{Data Type}                     & \textbf{Categories}      & \textbf{Train instances} & \textbf{Test instances} \\ \hline \hline
\multirow{2}{*}{Segmentation} 
                              & Pig Liver       & 50149                    & 5520                    \\
                              & Pig Gallbladder & 49812                    & 5520                    \\\cline{2-3} \hline
\multirow{2}{*}{Keypoints}   
                              & FBF             & 5476                     & 1372                    \\
                              & PCH             & 7320                    & 1831                    \\\cline{2-3}
\hline
\hline
\end{tabular}
}
\caption{Description of the annotated dataset. Keypoints were annotated for the Fenestrated Bipolar Forceps (FBF) and Permanent Cautery Hook (PCH). Train/Test instances refer to Section~\ref{Training_seg}.}
\label{tab:customdata_seg}
\end{table}

\subsubsection{Custom Segmentation Dataset}

In our previous work~\cite{oh2023framework}, a custom dataset of annotated images of a pig's liver and gallbladder was created and used to train the Detectron2~\cite{wu2019detectron2} object segmentation model. This dataset shares characteristics with the dataset described by Colleoni \textit{et al.}~\cite{colleoni2020synthetic}, where the robot arms and endoscope were manually maneuvered around the object without performing any actions on the tissues. Furthermore, the dataset in Oh \textit{et al.}~\cite{oh2023framework} was limited to a single liver sample, which presented significant challenges when applied to other samples or when the shapes and colors of tissues changed due to energy delivery. Consequently, the model trained on this dataset struggled with real-time tissue recognition during automated procedures. Furthermore, the dataset's size was relatively small compared to modern datasets, as video frames were down-sampled, and each frame had to be manually annotated.

\begin{figure}[!ht]
\centering
\includegraphics[width=\linewidth]{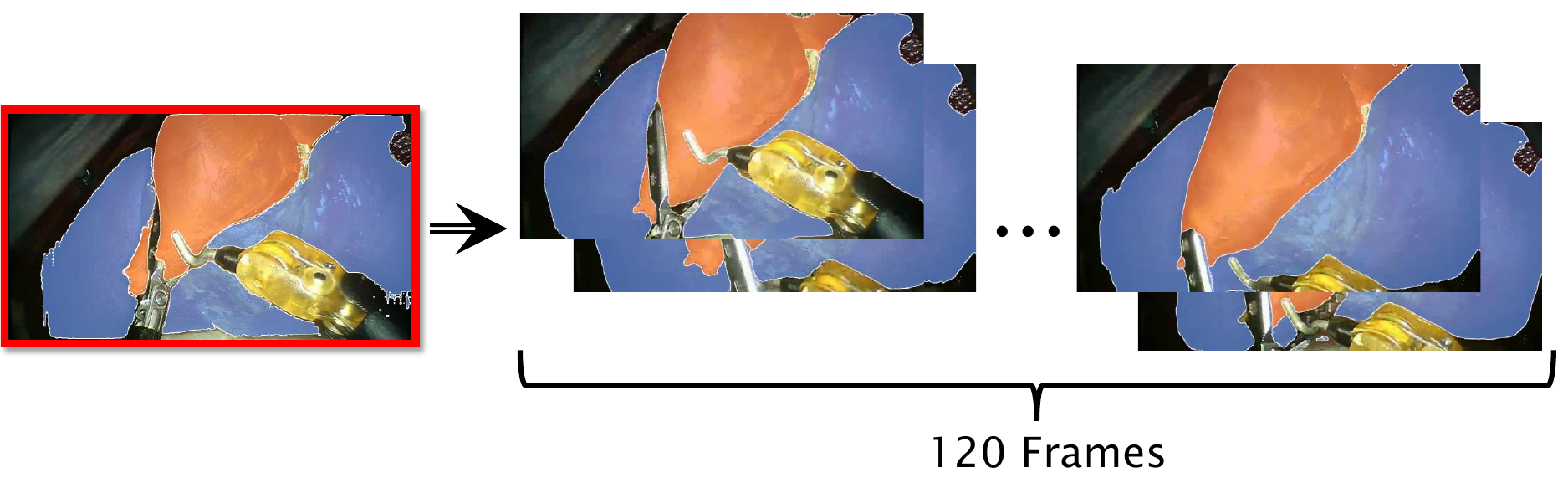}
\caption{An example of generating annotations with Track-Anything. Once the initial frame of the video clip (red box) is annotated, Track-Anything starts annotating the rest of the frames.}
\label{fig:trackany}
\end{figure}

We address these limitations by creating a new dataset annotated by Track Anything (TA)~\cite{yang2023track}. This new dataset utilized surgical videos from two different surgeons (E and F), one depicting an almost ideal cholecystectomy and the other showcasing a procedure in a challenging surgical environment. This choice was made to represent diverse surgical scenarios, ensuring that complex situations encountered in actual surgeries can be handled accurately. A limitation of TA is its reduced computational efficiency with the increasing number of video frames. To mitigate this, we split the videos into short clips lasting $2$ seconds (equivalent to $120$ frames).
After annotating the first frame, TA automatically extended the annotations to the remaining frames (Fig.~\ref{fig:trackany}). The variance within the video is minimal, except when the user moves the instruments rapidly. In such cases, we can reannotate the frame where the variation starts, and TA will update the remaining frames accordingly. 
The training set ultimately included around $55,000$ annotated images, which is approximately 35 times the size of the dataset from our previous work~\cite{oh2023framework}.

\subsubsection{Custom Keypoint Dataset}

\begin{figure}[!ht]
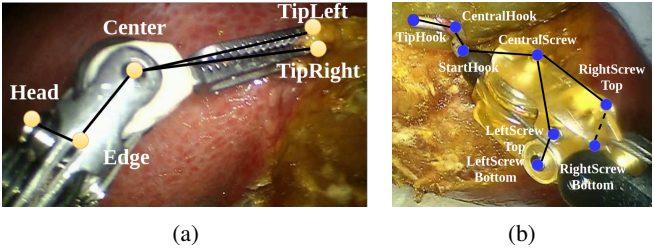

    \begin{subfigure}[b]{0.565\linewidth}
        \centering
        \includegraphics[width=\linewidth]{figures/keypoints_structure_FBF.png}
        \caption{}
        \label{fig:kpt_ann_fbf}
    \end{subfigure}
    \hfill
    \begin{subfigure}[b]{0.4\linewidth}
        \centering
        \includegraphics[width=\linewidth]{figures/keypoints_structure_PCH.png}
        \caption{}
        \label{fig:kpt_ann_pch}
    \end{subfigure}
\caption{KeyPoints structure for both the FBF and the PCH tools.}
\label{fig:kpt_anns}
\vspace{-3mm}
\end{figure}

\begin{figure}[!ht]
    \centering
    \includegraphics[width=0.7\linewidth]{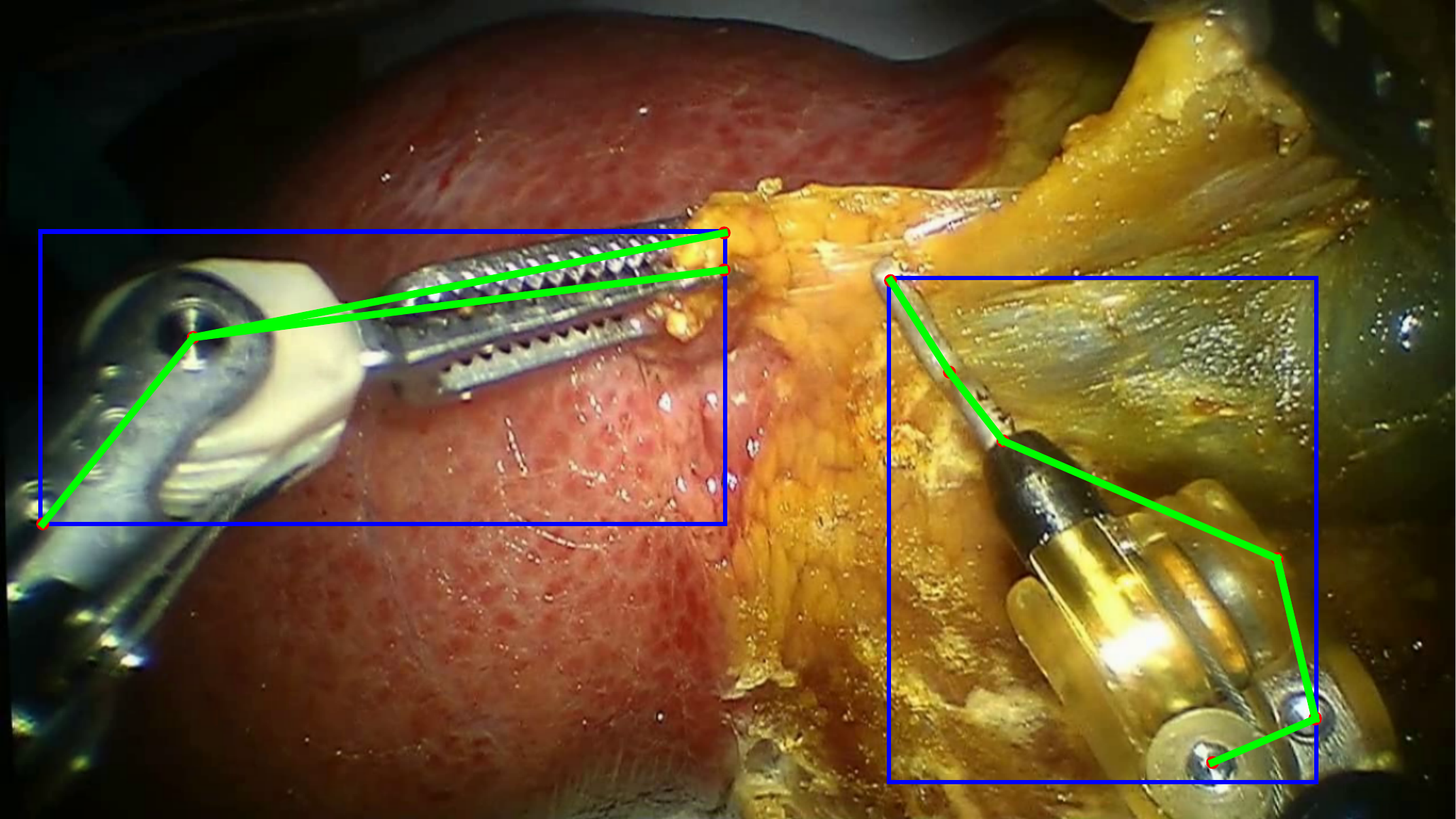}
\caption{Example of keypoint predictions by the trained Detectron2 model.}
\label{fig:kpt_preds}
\vspace{-3mm}
\end{figure}

Keypoint annotations of instruments (Large Needle Driver (LND), Fenestrated Bipolar Forceps (FBF), and Permanent Cautery Hook (PCH)) were performed manually using the COCO annotator~\cite{cocoannotator}. The keypoints selected for each instrument capture discriminative features and ensure consistency, as they should remain invariant to common transformations. The selected points are shown in Fig.~\ref{fig:kpt_anns}, and an example of the keypoint predictions is shown in Fig.~\ref{fig:kpt_preds}. The keypoints for LND are identical to our previous work~\cite{oh2023framework}.

These keypoints are on parts of the instruments with distinct colors and edges to maximize their detection. Details of the number of instances annotated for segmentation and keypoint detection are described in Table~\ref{tab:customdata_seg}. The dataset adheres to Microsoft’s COCO format~\cite{mscoco}, ensuring compatibility and ease of integration.

\section{Surgical Task} \label{section:task}

\subsection{Setup}

\begin{figure}[!ht]
    \centering
    \includegraphics[width=\linewidth]{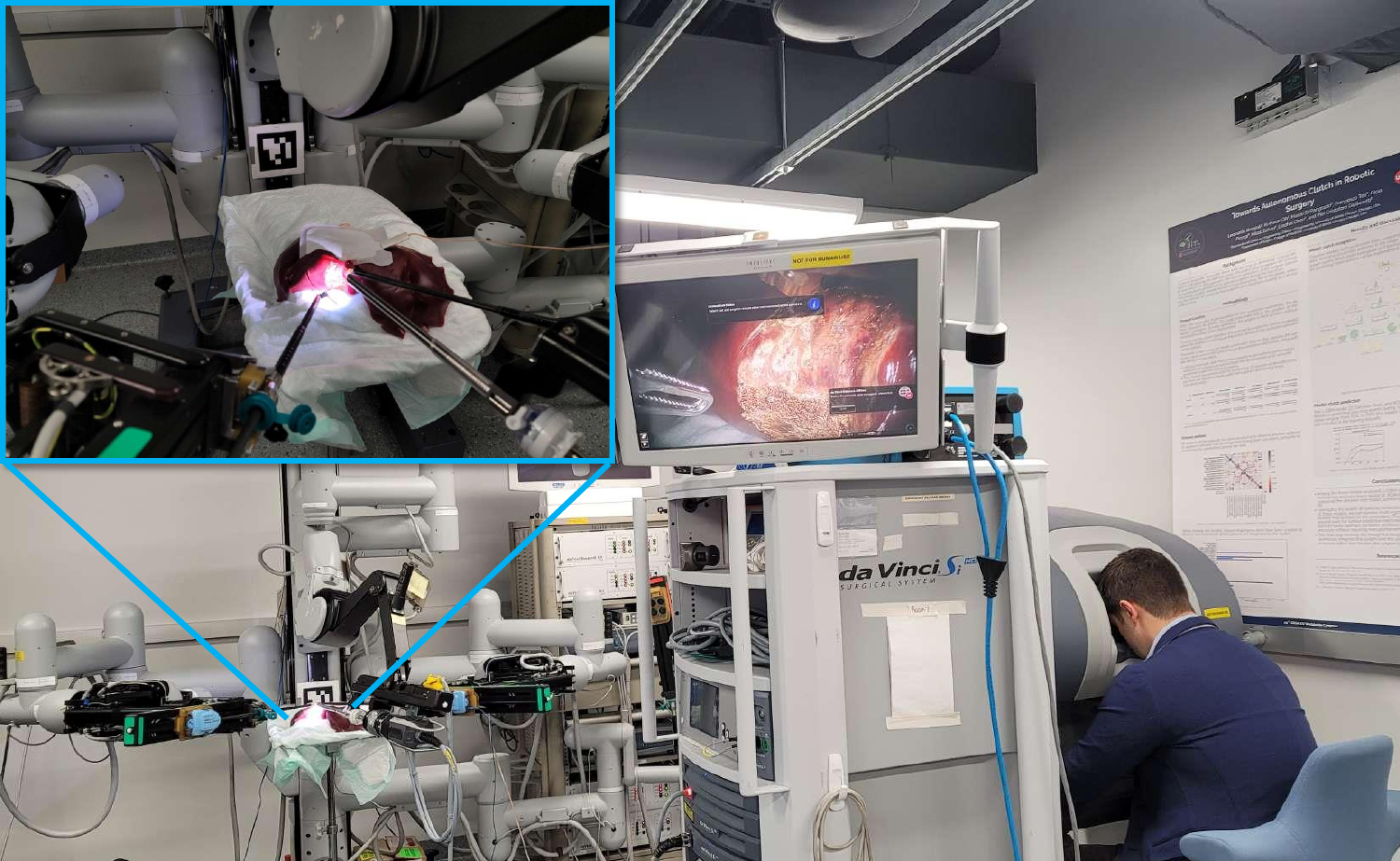}
\caption{The environment setup for the ex-vivo cholecystectomy performed by a surgeon.}
\label{fig:envsetup}
\end{figure}

The recordings took place in the setup depicted in Fig.~\ref{fig:envsetup}, where the surgeon controls the robot with the da Vinci console and executes the assigned task on a pig liver placed freely on a table with the gallbladder covered by the liver requiring assistance to lift it; this closely matches the actual in-vivo procedure. The dataset comprises seven surgeons denoted alphabetically from ``A'' to ``G'' who all have experience in surgical robotic cholecystectomy. Each subject performed the task three times, using a new liver for each attempt. The duration of the task varied according to the difficulty level of the task, influenced by factors such as the decay of the liver. In particular, challenges arose when the liver and the gallbladder were similar in color, making it difficult to distinguish between the two organs. 

Table~\ref{tab:surgdata} provides details on the data recorded for each surgeon. Note that some videos were damaged during compression and were consequently excluded from the dataset. In addition, occasional shutdowns of the Arduino occurred when a high current was applied to the tip of the instrument, resulting in corruption of pedal recordings.

\begin{table}[!ht]
\centering
\resizebox{\columnwidth}{!}{%
\begin{tabular}{c|c|c|c|c}
\hline
\hline
         \textbf{Surgeon}    & \textbf{Video} & \textbf{Kinematics} & \textbf{Pedals} &\textbf{Experience}\\
         &&&&(\# Procedures) \\ \hline \hline
{\color{red}\textbf{A}}    & \color{red}1             & \color{red}3                  & \color{red}3    & \color{red}150         \\
\textbf{B}    & 3             & 3                  & 3 & 1500              \\
{\textbf{C}}    & 3             & 3                  & 3  & 225            \\
{\color{red}\textbf{D}}    & \color{red}0             & \color{red}3                  & \color{red}3 & \color{red}65               \\
\textbf{E}    & 3             & 3                  & 3         & 1000     \\
{\color{red}\textbf{F}}    & \color{red}3             & \color{red}3                  & \color{red}0       & \color{red}225        \\
\textbf{G}    & 3             & 3                  & 3        & 1000      \\ \hline

\textbf{Total} & 16            & 21                 & 18        & -     \\ 
\hline
\hline
\end{tabular}}
\caption{Contribution of each surgeon to the dataset. Some recordings are excluded due to corruption.}
\label{tab:surgdata}
\vspace{-5mm}
\end{table}

\subsection{Procedure} \label{section:task_desc}
The surgeons performed the task following the UIC standardized surgical technique for robotic cholecystectomy~\cite{giulianotti2024foundation}. It should be noted that the order of certain steps can potentially vary depending on the specific surgical case or anatomical considerations. The primary steps of the procedure are as follows:

\begin{enumerate}
    \item Working area exposure
    \item Gallbladder neck retraction
    \item Calot triangle: anterior peritoneal layer opening
    \item Calot triangle: posterior peritoneal layer opening
    \item Cystic duct isolation
    \item Cystic artery isolation
    \item Cystic duct clipping
    \item Cystic artery clipping
    \item Cystic duct and artery division
    \item Detachment of the gallbladder from the liver
    \item Specimen retrieval in an Endobag\textsuperscript{TM}
\end{enumerate}

However, certain simplifications were applied to the technique mentioned above for this study and within the context of this experimental animal model. In particular, steps 8, 9, and 11 were omitted.

\subsection{Surgeon Profiles}

Understanding the backgrounds and experiences of surgeons is critical for analyzing RAS data. The skills and expertise of surgeons significantly impact the successful execution of robotic procedures, given the nuanced control and precise manipulation required. In our expanded dataset, we provide the surgical background of each participant. The dataset includes a table reporting the total number of procedures, the number of laparoscopic and robotic cases, and the hours of training in robotic surgery. A procedure is counted if it is endoscopically guided. Moreover, the procedures are categorized based on their complexity (low/mid/high). Several studies~\cite{lap_rob_learning_curve,lap_rob_learning_curve_2} demonstrate a direct connection between laparoscopic and robotic-assisted surgical skills, highlighting the relevance of this information. The total experience is shown in Table~\ref{tab:surgdata}, and more details are available in our GitHub repository.

This information is essential for developing models that predict a surgeon's performance and optimize the robotic system's assistance during surgery. By incorporating these surgeon profiles, we aim to enhance the personalization and effectiveness of robotic surgical systems, ultimately improving surgical outcomes.

\section{Preliminary Work}

This section aims to present related work that highlights the utility of different components offered by our dataset. Our goal is to showcase the validity and relevance of the various elements of the dataset, such as the efficiency of the keypoint tracking and the segmentation of
images.


\subsection{Pedal Intent Recognition}

In the context of robotic cholecystectomy, recognizing the intent of the surgeon's actions, particularly those that involve clutching and manipulating camera pedals, is essential to optimize procedural efficiency and alleviate the surgeon's cognitive workload. Upon activating the clutch, the orientations of both the robot arms (PSMs or ECM) and the manipulators (MTMs) are locked in place. However, during this state, the manipulators retain the ability to move while the positions of the da Vinci arms remain stationary.
In the current setting, the surgeon takes full control of the robotic system and receives no additional assistance. However, there is the potential to help the surgeon by automating some of the actions needed to operate the system effectively. For example, a dataset that includes the robot's kinematics and pedal signals can be used to develop an assistive system that could automatically activate the pedals. In our previous work~\cite{CRCD}, we described a preliminary version of a system that predicts when the clutch should be engaged. We next describe a much more robust and improved version of such a system below.

\subsubsection{Data processing}

The pedals and kinematic data were first synchronized due to different sampling rates. We adopted a sliding window-based approach for time series data~\cite{tonekaboni2021unsupervised}. The training data was generated by randomly sampling windows from the synchronized data. Thus, each training sample had the size $X \in \mathbb{R}^{f \times w}$, where $f$ is the number of features (the pose of each arm and manipulator), and $w$ is the size of the window.

These data were used to train the Time Series Transformer (TST)~\cite{tsai,zerveas2020transformerbasedframeworkmultivariatetime}. This model first encodes the time-series input samples to fit the input requirements of the original transformer encoder~\cite{vaswani2017transformers}. The output of the transformer encoder is then passed to a linear output layer, and the model is trained to minimize the squared error between the predictions and the ground truth labels.

\subsubsection{Training Results}

The model was trained using different window sizes, $w \in \{40, 60, 80\}$. Selecting the appropriate window size is crucial, impacting how quickly the model can make predictions in real-time applications. However, if the window size is too large, it may include redundant information or contain more than one pedal action within the window (e.g., pressing the clutch and camera pedals sequentially).
The composition of the dataset is shown in Table~\ref{tab:class_distribution}. Since the data for different classes are imbalanced (the pedals are mostly not pressed), we reduced the size of the majority class so that the ratio between the two classes was 15; we arrived at this ratio experimentally (see also~\cite{CRCD}). Subsequently, the data was split in a $7:3$ ratio for training and testing. Table~\ref{tab:evalmetrics} presents the precision, recall, and F1 score for each trained model tested on the test set.

\begin{table}[!ht]
\centering
\resizebox{0.7\columnwidth}{!}{%
\begin{tabular}{l|c|c}
\hline\hline
\textbf{Pedal Type} & \textbf{Not Pressed} & \textbf{Pressed}  \\ \hline \hline
\textbf{Clutch}     & 1082871              & 4845              \\ \hline
\textbf{Camera}     & 967704               & 31095             \\ \hline \hline
\end{tabular}
}
\caption{Composition of the pedal dataset.}
\label{tab:class_distribution}
\vspace{-3mm}
\end{table}

\begin{table}[!ht]
\centering
\resizebox{\columnwidth}{!}{%
\begin{tabular}{l|ccc|ccc}
\hline \hline
                                 & \multicolumn{3}{c|}{\textbf{Camera}} & \multicolumn{3}{c}{\textbf{Clutch}} \\ \hline \hline
\multicolumn{1}{l|}{\textbf{Window Size}} & 40      & 60      & 80      & 40      & 60      & 80     \\ \hline
\multicolumn{1}{l|}{\textbf{Precision}}   & 0.995   & 0.991   & 0.996   & 0.957   & 0.940   & 0.987  \\ \hline
\multicolumn{1}{l|}{\textbf{Recall}}      & 0.967   & 0.968   & 0.977   & 0.992   & 0.960   & 0.990  \\ \hline
\multicolumn{1}{l|}{\textbf{F1 Score}}    & 0.981   & 0.979   & 0.987   & 0.974   & 0.950   & 0.989  \\ \hline \hline
\end{tabular}
}
\caption{Accuracy, Recall, and F1 scores measured on the test set}
\label{tab:evalmetrics}
\vspace{-3mm}
\end{table}

\begin{figure}[!ht]
    \centering
    \includegraphics[width=\linewidth]{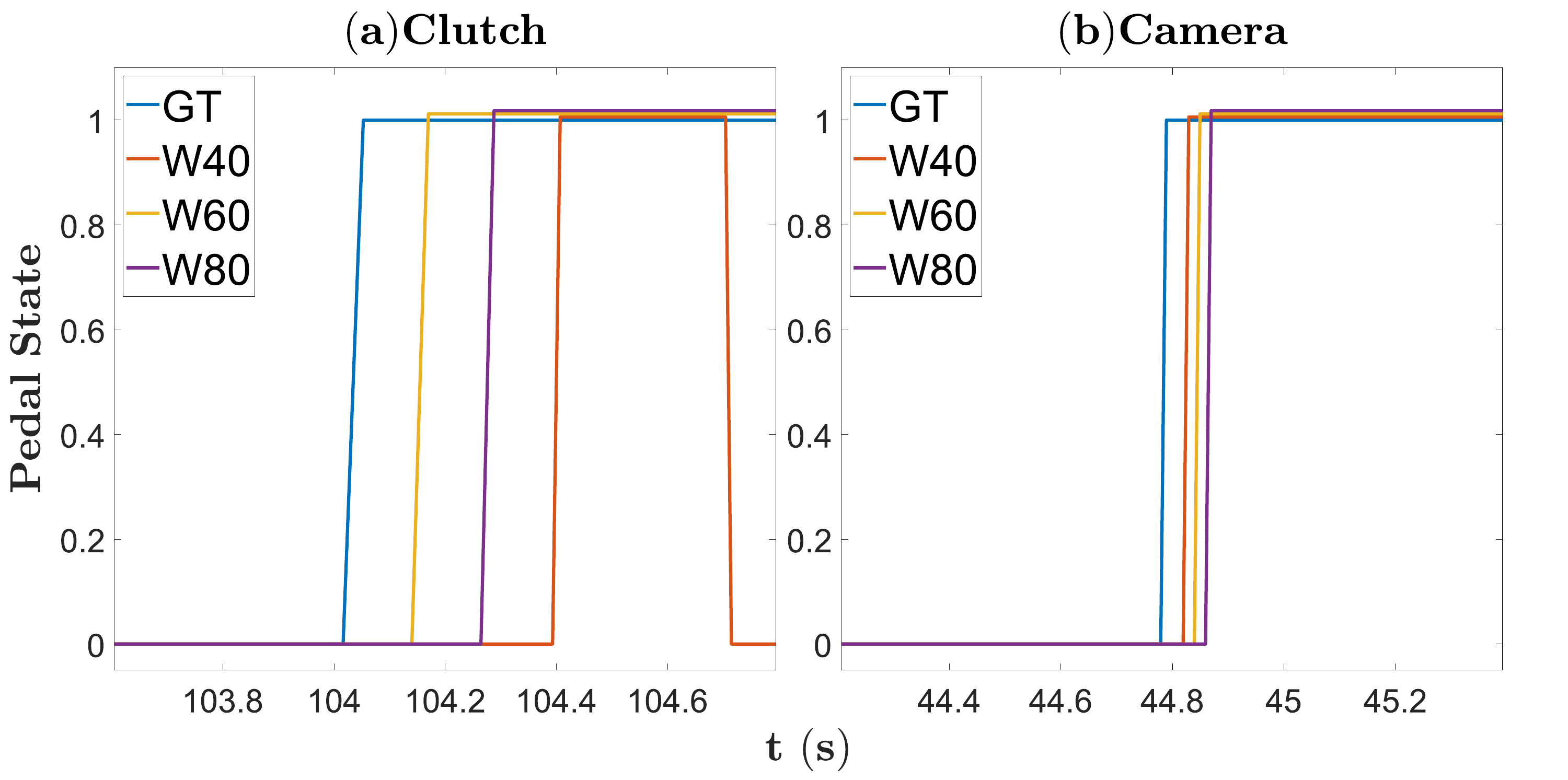}
    \caption {Zero-shot test of pedal prediction. The left figure shows the clutch pedal, while the one on the right shows the camera pedal prediction for three different windows (40, 60, 80).}
    \label{fig:predictions_graph}
    \vspace{-3mm}
\end{figure}

Fig.~\ref{fig:predictions_graph} shows the performance of the model on data from a surgeon whose recordings were not included in the original train/test data (zero-shot test). The trained TST models predicted the pedal states by sliding a window with a step size of 2 samples. The colors represent the models trained on different window sizes. The models performed similarly in predicting when to press the camera pedal. However, their performance differs for the clutch pedal. In that case, the model trained with window size 60 performs best. This suggests that window sizes larger than 60 lead to overfitting (Table~\ref{tab:evalmetrics} on its own is not sufficient to show this), while shorter windows fail to model the data correctly. The difference in performance on camera and clutch pedals is due to surgeons pressing the camera pedal more frequently than the clutch pedal, resulting in the sample imbalance shown in Table~\ref{tab:class_distribution}. Nevertheless, these results suggest that with enough data from diverse procedures, it is possible to train a highly accurate model.

\subsection{Instance Segmentation and Keypoint Detection}\label{Training_seg}

To automate aspects of robotic cholecystectomy, the robot must recognize and keep track of tissues during the procedure. Currently, there is a notable scarcity of datasets specifically designed for such research. However, in contrast to existing ones, this new dataset uniquely captures dynamic tissue changes during cholecystectomy procedures. In particular, tissues exhibit a rich diversity in both shape and color. This deliberate inclusion of diverse tissue characteristics is pivotal for training segmentation models, enabling the robot to recognize and track tissues in real-time during surgical procedures.

\subsubsection{Trained Models}

Detectron2~\cite{wu2019detectron2}, derived from Mask R-CNN~\cite{he2018maskrcnn}, has two independent types of models: instance object segmentation models and human pose estimation models (or keypoint detection models). Throughout this paper, we distinguish the two models as DT2-Seg and DT2-Kpt to avoid confusion. The hyperparameters used for training the models and further details can be found in our previous work~\cite{CRCD}.

In this paper, we add training results from MaskDINO~\cite{li2022mask}, a state-of-the-art object detection and segmentation model. For our work, we increased the base learning rate of MaskDINO to 0.0004 and decayed in steps 1000 and 2000 to avoid overfitting and train faster~\cite{li2022mask}. Moreover, we decreased the images per batch to 16 and the total number of iterations to 3000 since the dataset size is small compared to the original COCO~\cite{mscoco} dataset used to train MaskDINO. The rest of the parameters are set to default.


\subsubsection{Training Results}

We trained DT2-Seg and MaskDINO with the segmentation dataset and trained DT2-Kpt with the keypoint dataset as described in Table~\ref{tab:customdata_seg}. Subsequently, DT2-Seg and MaskDINO models were evaluated on an independent dataset of 5520 images from one of Participant C's videos. Table~\ref{tab:dt2res} compares the Average Precision (AP)~\cite{mscoco} results for different models on this independent dataset. 

\begin{table}[!ht]
\centering
\resizebox{\columnwidth}{!}{%
\begin{tabular}{l|l|c|c|c}
\hline \hline
                                         & \textbf{Categories} & \textbf{AP (Box)} & \textbf{AP (Seg.)} & \textbf{AP (Keypt.)} \\ \hline \hline
\multirow{2}{*}{\textbf{DT2-Seg}}        & Liver               & 63.2              & 68.4               & -                    \\
                                         & Gallbladder         & 68.3              & 67.8               & -                    \\ \hline
\multirow{2}{*}{\textbf{MaskDINO}}       & Liver               & 91.2              & 90.0               & -                    \\
                                         & Gallbladder         & 83.7              & 84.1               & -                    \\ \hline
\multirow{2}{*}{\textbf{DT2-Kpt}}        & FBF                 & 77.1              & -                  & 94.6                 \\
                                         & PCH                 & 74.2              & -                  & 98.4                 \\  \hline \hline
\end{tabular}}
\caption{The Average Precision (AP) scores (percentages) for each category (Box stands for Bounding Box and Seg. for Segmentation).}
\label{tab:dt2res}
\end{table}

\begin{figure}
\centering
    \begin{subfigure}[b]{0.48\linewidth}
        \centering
        \includegraphics[width=\linewidth]{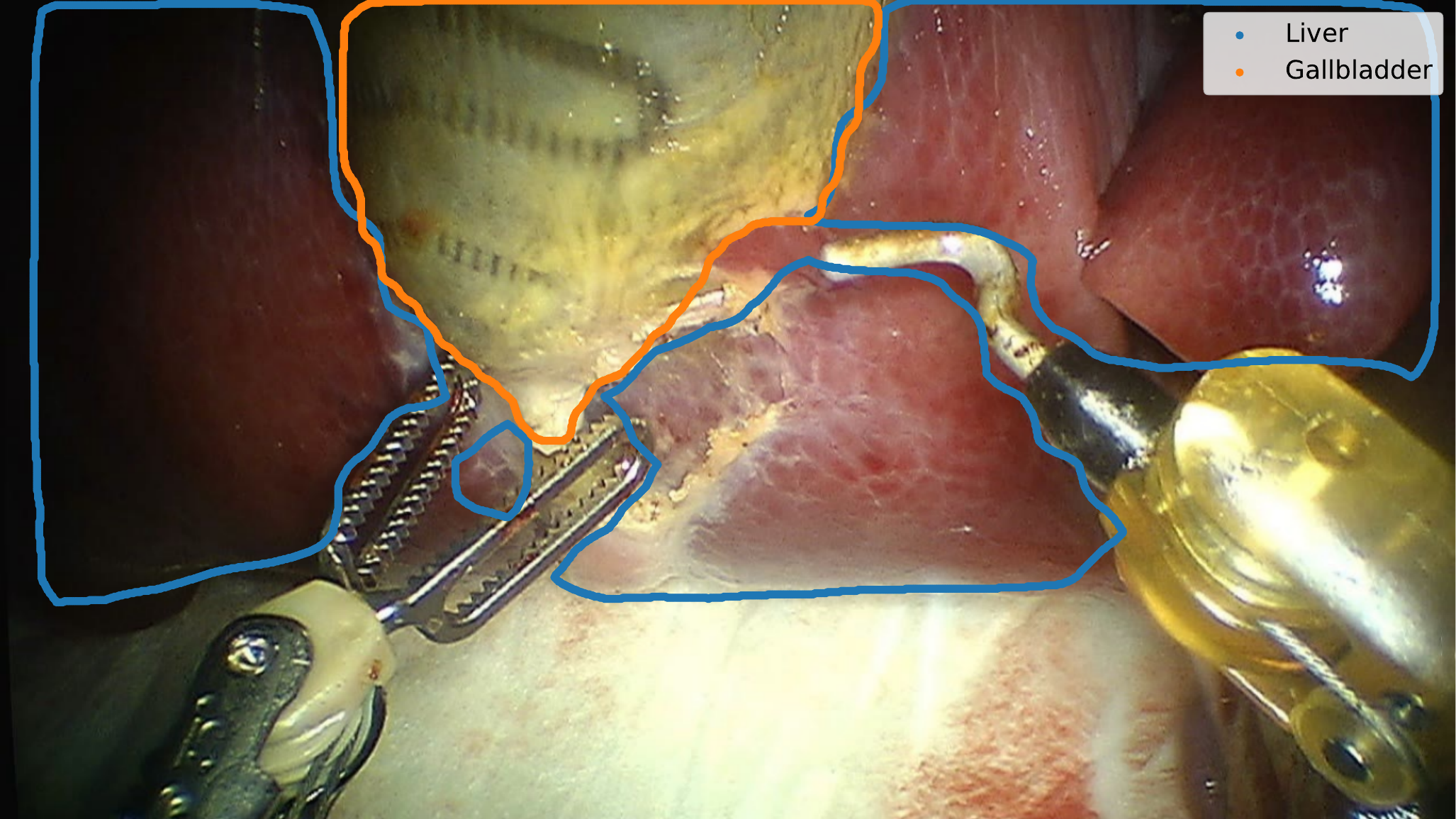}
        \caption{}
        \label{fig:dt2_pred}
    \end{subfigure}
    \hfill
    \begin{subfigure}[b]{0.48\linewidth}
        \centering
        \includegraphics[width=\linewidth]{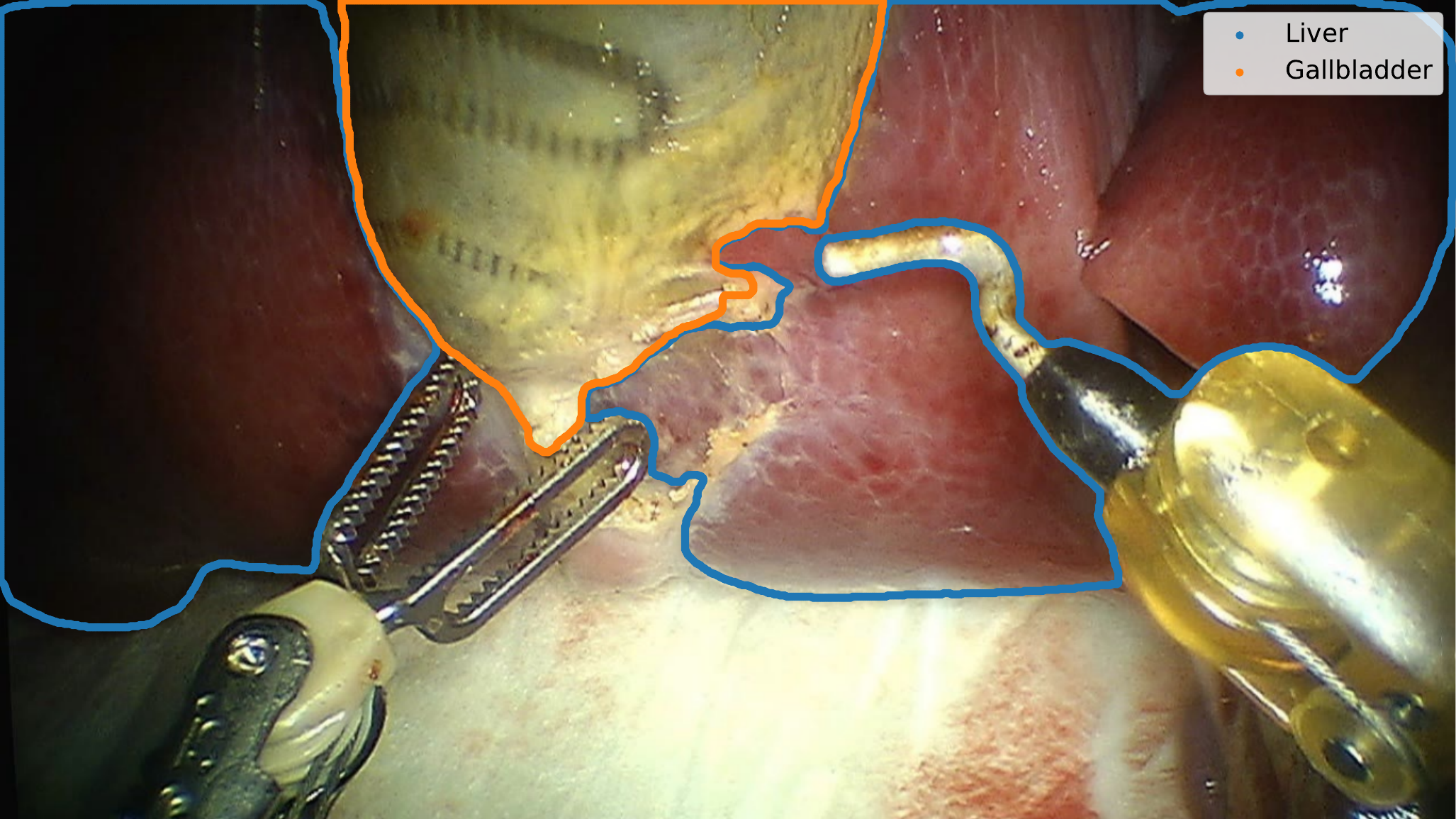}
        \caption{}
        \label{fig:maskdino_pred}
    \end{subfigure}
\caption{Segmentation prediction results using Detection2 (left) and MaskDino (right).}
\label{fig:seg_preds}
\vspace{-3mm}
\end{figure}

To further investigate the difference in performance, we applied the two models to an image in the video from surgeon E that was not included in the training set (Fig.~\ref{fig:seg_preds}). Both models managed to segment the different tissues. However, DT2-Seg had problems isolating the instruments from the tissues, while MaskDINO filtered the surgical instrument out precisely. Furthermore, for MaskDino, the identified boundaries between the two tissues were more accurate and less noisy.

\subsection{3D Scene Reconstruction}

A significant component of our dataset is the intrinsic and extrinsic parameters of the endoscope. These parameters were used, for example, in our previous work~\cite{oh2023framework} for 3D reconstruction of the surgical scene. The cameras were calibrated following the traditional approach outlined in~\cite{zhang2000} using MATLAB Stereo Camera Calibration Toolbox and OpenCV~\cite{opencv_library}.

\begin{figure}[!ht]
    \centering
    \includegraphics[trim=10cm 13cm 10cm 10cm, clip, width=\linewidth]{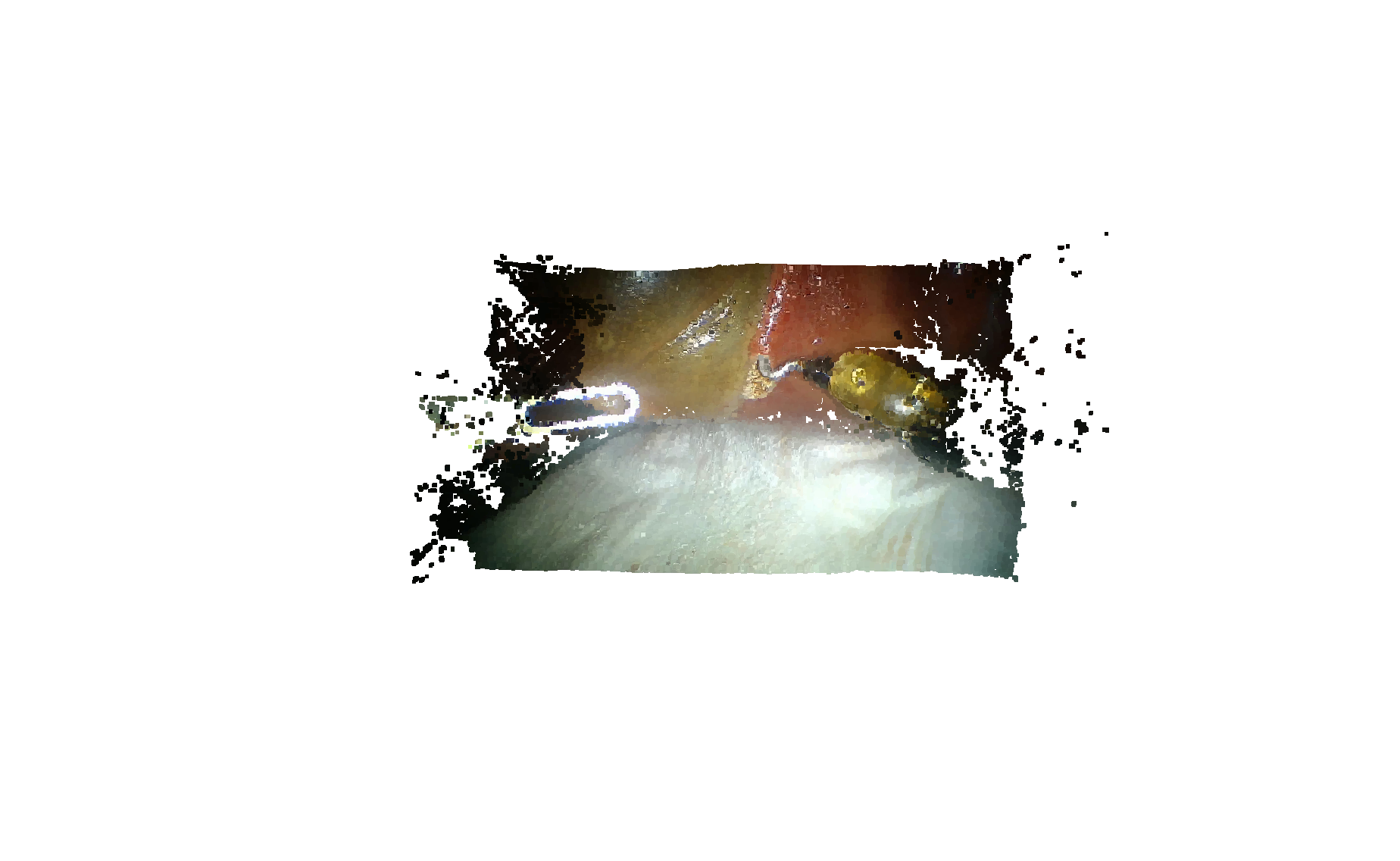}
    \caption{Generated point cloud from the images in Fig.~\ref{fig:stereo}}
    \label{fig:cloudpoint}
\end{figure}

Subsequently, using the intrinsic and extrinsic camera parameters, we applied the modified Semi-Global Matching algorithm (SGM)~\cite{sgm} to produce stereo disparity maps from stereo endoscopic images. Before applying the SGM, the images were passed through a bilateral filter to reduce noise while preserving edges as much as possible. The disparity map was then converted to point clouds using the baseline and focal length of the stereo cameras. The generated point clouds are dictionaries of the form $(\{(u, v)|(x, y, z)\})$, where the estimated 3D point $(x, y, z)$ is mapped to each pixel $(u, v)$ in the 2D image. Fig.~\ref{fig:cloudpoint} shows an example of the point cloud generated from the stereo endoscopic images in Fig.~\ref{fig:stereo}.

While the SGM~\cite{sgm} algorithm does not perform as well as state-of-the-art methods based on AI~\cite{scharstein2002taxonomy}, its performance was adequate for our work. Further, SGM~\cite{sgm} does not require high-end hardware to run in real-time.

\begin{table*}[!t]
\centering
\resizebox{\textwidth}{!}{%
\begin{tabular}{c|c|c|c|c|c}
\hline \hline
\multirow{2}{*}{\textbf{Name}} & \multirow{2}{*}{\textbf{Year}} & \multirow{2}{*}{\textbf{Data}} & \multirow{2}{*}{\textbf{Procedure}} & \multirow{2}{*}{\textbf{Annotations}} & \multirow{2}{*}{}\textbf{Annotated}  \\
&&&&&\textbf{Frames}\\

\hline \hline
JIGSAWS ~\cite{gao2014jhu} & 2014 & 103 videos + kinematic data & In-vitro experiments  & Gestures, scoring & - \\

\hline
EndoVis 2017 ~\cite{murali2023endoscapes}& 2017 & 8 videos & Porcine procedures  & Tool segmentation & 3000 \\
\hline
FlapNet ~\cite{attanasio2020autonomous}& 2020 & 1 video & Lobectomy  & Tissue flap and tools& 2160\\
\hline
UCL dVRK~\cite{colleoni2020synthetic} & 2020 & 14 videos + kinematic data & Synthetic background  & Tool segmentation &4200 \\ 
\hline
RoboTool~\cite{garcia2021image} & 2021 & 20 videos & Different & Tool segmentation& 514\\
\hline
AutoLaparo ~\cite{wang2022autolaparo} & 2022 & 21 videos & Hysterectomy &  Tasks perception & 5936\\
\hline
Hemoset ~\cite{HemoSet}& 2024 & 11 videos & Thyroidectomy &Blood Segmentation &857\\
\hline
\multirow{2}{*}{CRCD ~\cite{CRCD}}  & \multirow{2}{*}{2024} & \multirow{2}{*}{}16 videos + kinematic data  & \multirow{2}{*}{Cholecystectomy} & \multirow{2}{*}{-}  & \multirow{2}{*}{-} \\
&& + pedals &&&\\
\hline
\multirow{2}{*}{Exd. CRCD}  & \multirow{2}{*}{2024} & \multirow{2}{*}{}16 videos + kinematic data & \multirow{2}{*}{Cholecystectomy} &  \multirow{2}{*}{}Tool keypoints, & \multirow{2}{*}{127000} \\
&& + pedals &&Tissue segmentation&\\
\hline\hline
\end{tabular}
}
\caption{Public available datasets in Robotic-assisted surgery.}
\label{tab:compariason_table}
\vspace{-3mm}
\end{table*}

\section{Dataset Comparison}

We compared our expanded version of the CRCD~\cite{CRCD} dataset with publicly available surgical robotics datasets~\cite{dataset_review,rueckert2024methods}. A dataset is only considered if the robotic surgical instruments are visible in the videos and it offers segmentation or kinematic ground truth data. The comparison is shown in Table~\ref{tab:compariason_table}.

Apart from our dataset, only two datasets~\cite{gao2014jhu, colleoni2020synthetic} include kinematic data. However, JIGSAWS~\cite{gao2014jhu} does not contain any information on ECM, and the procedures are in-vitro, while UCL dVRK~\cite{colleoni2020synthetic} does not include information on MTMs. Neither dataset contains the pedal data. This underscores the unique contribution of our dataset, which provides the kinematics of all da Vinci arms and console manipulators, including pedal usage. Furthermore, in a typical dataset, the number of videos does not directly correspond to the number of annotated frames, as most available datasets annotate only a small subset of frames. In contrast, we annotated all frames in certain videos, providing at least an order of magnitude more annotated frames.

\section{Conclusion}

Most current applications of machine learning to RAS rely on annotated videos from existing well-known datasets. However, a notable gap exists due to the absence of kinematic data in these datasets. Other challenges persist, including incomplete recordings, lack of context awareness, imprecise kinematic data due to calibration issues, and reliance on artificial exercise-based scenarios rather than actual procedures.

To address these limitations, we introduced CRCD~\cite{CRCD}, a comprehensive dataset recorded during actual robotic cholecystectomy procedures on ex-vivo porcine livers. The expanded version of the dataset described in this paper includes not only patient-side kinematic data, pedal states, and timestamped videos but also information on the experience of participating surgeons, including data on the number and complexity of laparoscopic procedures, as well as their hours of training in robotic surgery. A complementary dataset of liver segmentations and keypoint annotations for tracking surgical instruments is also included. By integrating these elements, our dataset provides a richer context for surgical actions, allowing for more nuanced analysis and model training. A comparison of CRCD with the existing datasets is provided to highlight its limitations and clearly identify its main contributions.

To demonstrate the usefulness of the expanded dataset, we studied the performance of segmentation models, which is crucial for the robot's ability to recognize and track tissues during cholecystectomy. The results underscore the usefulness of our dataset in enhancing the robot's tissue recognition capabilities. The dynamic changes in the tissues during cholecystectomy procedures captured by the dataset contribute to improved models for real-time tissue recognition. We also show how our dataset can be used to train models for tracking surgical instruments, which are critical for the autonomous control of the robot. Further, expanding our previous work, we proposed a novel classifier to predict clutch and pedal usage. These applications show that the expanded dataset provides an important new resource for advancing automation in robotic cholecystectomy. Furthermore, by including detailed information on each surgeon's prior experience with RAS, our dataset can be used to investigate what level of assistance should be provided to the surgeon. Such assistance would alleviate the stress and burden on surgeons during prolonged surgical interventions, contributing to better surgical outcomes.

Our goal in creating CRCD is to provide a comprehensive public dataset, capturing all available signals from both the console and patient-side arms during surgeries performed by experts on porcine livers.
One of the main differences between in-vivo and ex-vivo procedures on porcine specimens is the endoscopic view. In in-vivo cases, the surgical field is generally brighter due to light reflections from the body wall. However, this does not affect tissue segmentation or instrument keypoint detection, as these elements are typically centered in the endoscopic view. From a surgical perspective, there is no significant difference: cholecystectomy is relatively simple due to the ample field of view at the surgical site. The liver was positioned as it would be in an actual procedure, with the gallbladder covered by the liver, requiring assistance to lift it.

The main limitation of our dataset derives from the difference in the size of the workspace between ex-vivo and in-vivo environments. The workspace of the arms is less constrained in an ex-vivo setting since the body wall is not present. In addition, in vivo procedures within body cavities are challenging for dVRK setup.

Combining video recordings, kinematic data, pedal signals, comprehensive annotations, and detailed surgeon profiles, CRCD has the potential to advance the field of robotic-assisted surgery significantly. By providing researchers with extensive and detailed information, our dataset allows the development of sophisticated models to improve the understanding and automation of surgical procedures, ultimately enhancing patient care.

\newpage
\clearpage
\nocite{*}
\bibliographystyle{IEEEtran}
\bibliography{jmrr_2024}

\vspace{10mm}

\noindent\includegraphics[width=1in]{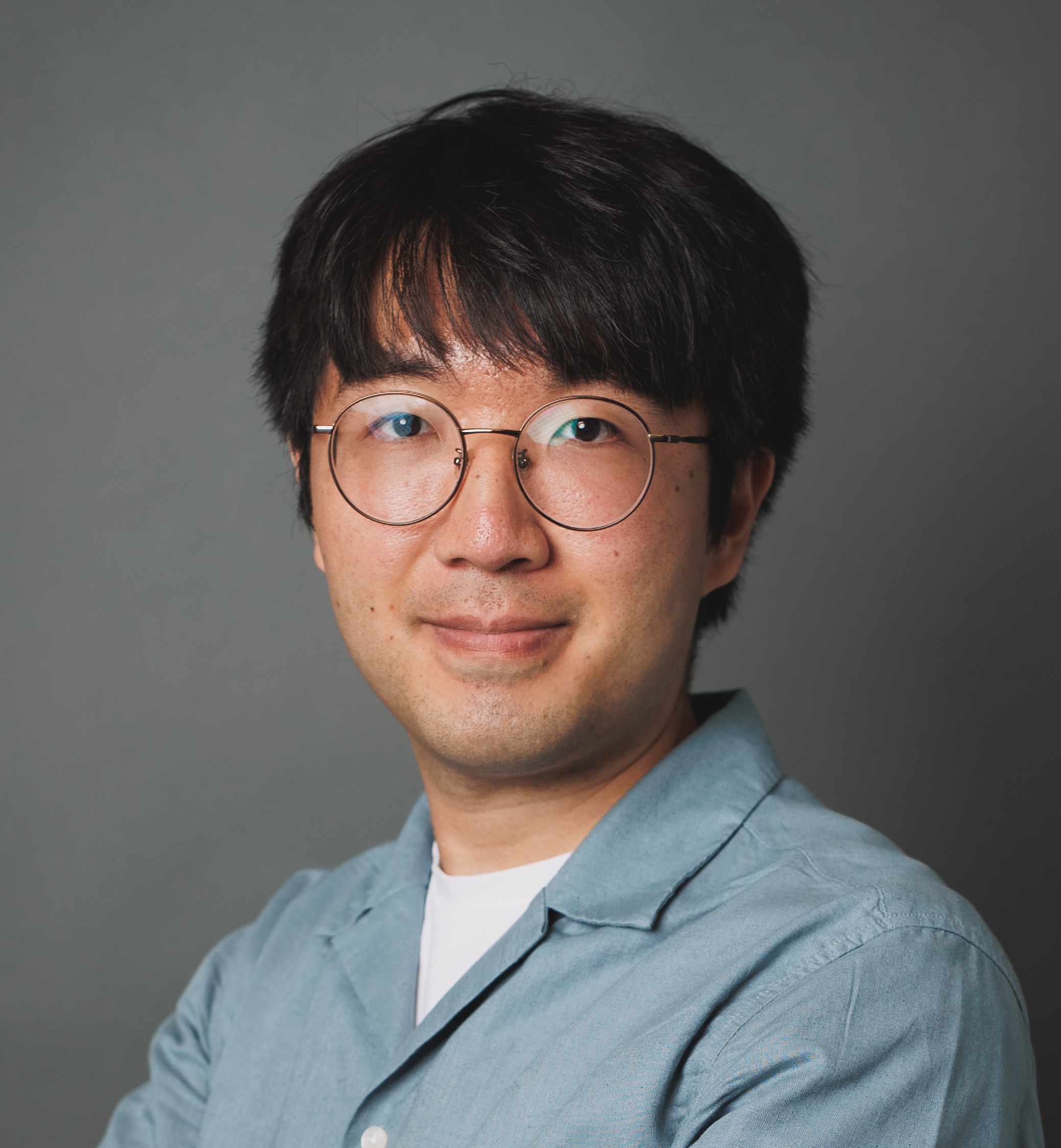}
{\bf Ki-Hwan Oh} received a B.S. degree in Electronic and Electrical Engineering from Sungkyunkwan University, Suwon, South Korea. He is working toward a Ph.D. in Electrical and Computer Engineering at the University of Illinois Chicago, Chicago, IL, USA. 
He is also a Graduate Research Assistant in the Surgical Innovation Training Laboratory (SITL), Department of Surgery, University of Illinois Chicago, Chicago, IL, USA. His research interests include modeling human-human interactions and automation of surgical robots. 

\vspace{10mm}

\noindent\includegraphics[width=1in]{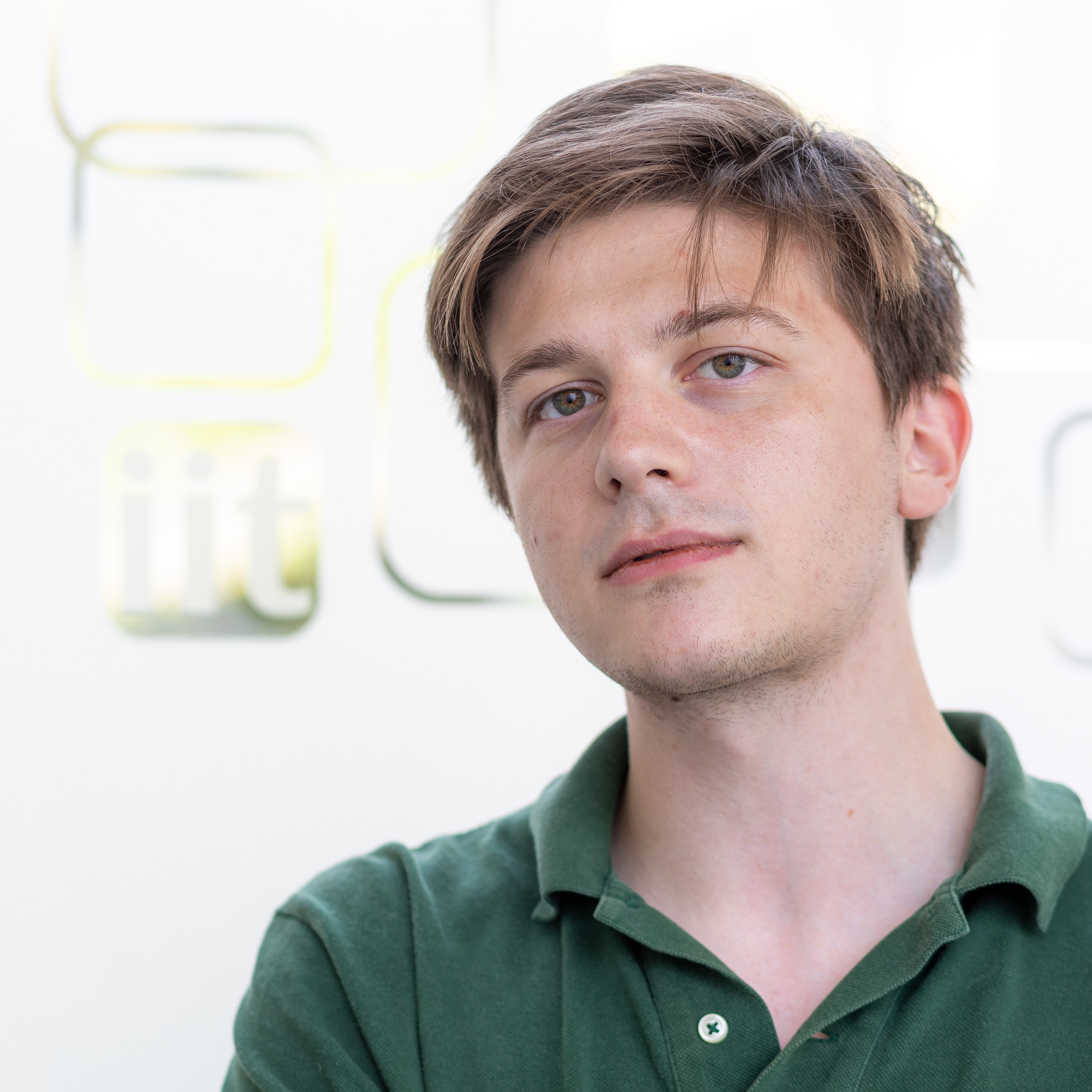}
{\bf Leonardo Borgioli} received his double M.S. degree in Advanced Robotics from the University of Genova, Italy, and Ecole Centrale Nantes, France. He is working toward a Ph.D. in Electrical and Computer Engineering at the University of Illinois Chicago, Chicago, IL, USA. He is also a Graduate Research Assistant in the Surgical Innovation Training Laboratory (SITL), Department of Surgery, University of Illinois Chicago, Chicago, IL, USA. His research interests include robotic-assisted surgery, specifically focusing on the application of artificial intelligence and virtual reality in this field.

\end{document}